\documentclass[conference]{IEEEtran}
\IEEEoverridecommandlockouts
\usepackage{cite}
\usepackage{amsmath,amssymb,amsfonts}
\usepackage{algorithmic}
\usepackage{graphicx}
\usepackage{textcomp}
\usepackage{nccmath} 
\usepackage{xcolor}
\usepackage{stfloats}
\usepackage{url}
\usepackage{makecell} 
\usepackage{booktabs}
\usepackage{cite}
\usepackage{colortbl} 
\usepackage{color}
\usepackage{multirow}
\usepackage[pdftex,linkcolor=blue,citecolor=blue,backref=page]{hyperref}

\def\BibTeX{{\rm B\kern-.05em{\sc i\kern-.025em b}\kern-.08em
    T\kern-.1667em\lower.7ex\hbox{E}\kern-.125emX}}

\title{Training-Free BioNEL-RAG: Advancing Biomedical Entity Linking through Retrieval-Augmented Generation and Large Language Models\\
}

\author{\IEEEauthorblockN{1\textsuperscript{st} Ziyi Chen}
\IEEEauthorblockA{\textit{School of Biomedical Engineering,}\\\textit{ Division of Life Sciences and Medicine,}\\\textit{ University of Science and Technology of China, }\\\textit{ Hefei, China} \\
}
\and
\IEEEauthorblockN{1\textsuperscript{st} Hui Liu}
\IEEEauthorblockA{\textit{Department of Precision Machinery and Precision Instruments,}\\\textit{ University of Science and Technology of China,}\\\textit{ Hefei, China}
}
\and
\IEEEauthorblockN{2\textsuperscript{nd} Ziyang Ling}
\IEEEauthorblockA{\textit{School of Biomedical Engineering,}\\\textit{ Division of Life Sciences and Medicine,}\\\textit{ USTC,  Hefei, China} 
}
\and
\IEEEauthorblockN{3\textsuperscript{rd} Ronald X. Xu\textsuperscript{*}}
\IEEEauthorblockA{\textit{School of Biomedical Engineering,}\\\textit{ Division of Life Sciences and Medicine,}\\\textit{ USTC,  Hefei, China} 
}
\and
\IEEEauthorblockN{4\textsuperscript{th} Mingzhai Sun\textsuperscript{*}\thanks{* Corresponding author: Ronald X. Xu (xux@ustc.edu.cn), Mingzhai Sun (mingzhai@mail.ustc.edu.cn). USTC: University of Science and Technology of China}}
\IEEEauthorblockA{\textit{School of Biomedical Engineering,}\\\textit{ Division of Life Sciences and Medicine,}\\\textit{USTC,  Hefei, China} 
}}

\begin{document}
\maketitle

\begin{abstract}
Biomedical Named Entity Linking (BioNEL) is the task of mapping mentions extracted from biomedical texts to canonical entities or concepts in the standardized knowledge base. Although existing methods have achieved remarkable performance in mapping accuracy, most rely on training with domain-specific datasets, which fundamentally limits their cross-domain adaptability. To address this limitation, we propose BioNEL-RAG, a novel training-free framework that combines retrieval-augmented generation (RAG) with large language models (LLMs). Within this framework, the RAG module focuses on retrieving candidate entities, while the large language model module plays the role of selecting the most appropriate candidate for entity normalization, which is based on the analysis of the mention's context, with the aid of its remarkable text comprehension capabilities. Experimental results demonstrate that BioNEL-RAG achieves performance comparable to existing methods without requiring training on subdomain datasets, whereas these methods necessitate fine-tuning on specific subdomains such as genes, diseases, or chemical compounds.
%Experimental results on datasets demonstrate that BioNEL-RAG, relying solely on models pretrained in the broader biomedical domain, achieves performance comparable to current methods that require fine-tuning on specific subdomains such as genes, diseases, or chemical compounds.
\end{abstract}


\section{Introduction}
%% 提出当前深度学习模型泛化能力不足，引出模型泛化性的重要性
Named Entity Linking (NEL), or normalization, is a fundamental natural language processing (NLP) task that maps textual mentions to unique identifiers in a knowledge base, typically following Named Entity Recognition (NER). Biomedical Named Entity Linking (BioNEL) is a specialized subset of NEL. In the biomedical domain, many deep learning models, particularly those BERT-based, have achieved high accuracy in addressing the aforementioned challenges, with performance in subdomains like diseases and chemical compounds often exceeding 93\%, leaving limited room for improvement through further architectural modifications\cite{sung-etal-2020-biomedical}. Although these models perform well within their respective subdomains, their effectiveness often declines when applied across domains. 

%% 引出大模型的泛化性很强，可以借用大模型的泛化能力，同时大模型又具有幻觉，所以引出RAG
Over the past two years, LLMs have attracted considerable attention for their strong generalization and instruction-following capabilities across a wide range of tasks. It show promise for overcoming the generalization limits of traditional deep learning in biomedical NEL. However, their reliance on general-domain data makes them prone to hallucination in specialized fields like biomedicine. While domain-specific fine-tuning can mitigate this, the process requires substantial data and computational resources, and the resulting model's generalization is often confined to its training dataset.

To address the aforementioned challenges, we propose BioNEL-RAG, a retrieval-augmented biomedical entity linking framework built upon an expandable knowledge base. The framework operates in two stages: it first retrieves top-n candidate entities for a target mention, then prompts an LLM to analyze the context and select the most appropriate one. The novelty of our work is that BioNEL-RAG achieves state-of-the-art (SOTA) performance without requiring task-specific training. Furthermore, its scalable architecture seamlessly adapts to updates from biomedical resources like UMLS\cite{bodenreider2004unified}, overcoming a key limitation of traditional and fine-tuned LLMs that struggle with evolving knowledge.
%BioNEL-RAG achieves state-of-the-art (SOTA) performance without requiring task-specific training, unlike prior models. Furthermore, its scalable architecture seamlessly adapts to updates from biomedical resources like UMLS\cite{bodenreider2004unified}, overcoming a key limitation of traditional and fine-tuned LLMs that struggle with evolving knowledge.

%%%%%%%%%%%%%%%%%%%%%%%%%%%%%%%%%%%%%%%%%%%%%%%%%%%%%%%%%%%%

\begin{figure*}[htb]
\centering
\includegraphics[width=1.0\textwidth]{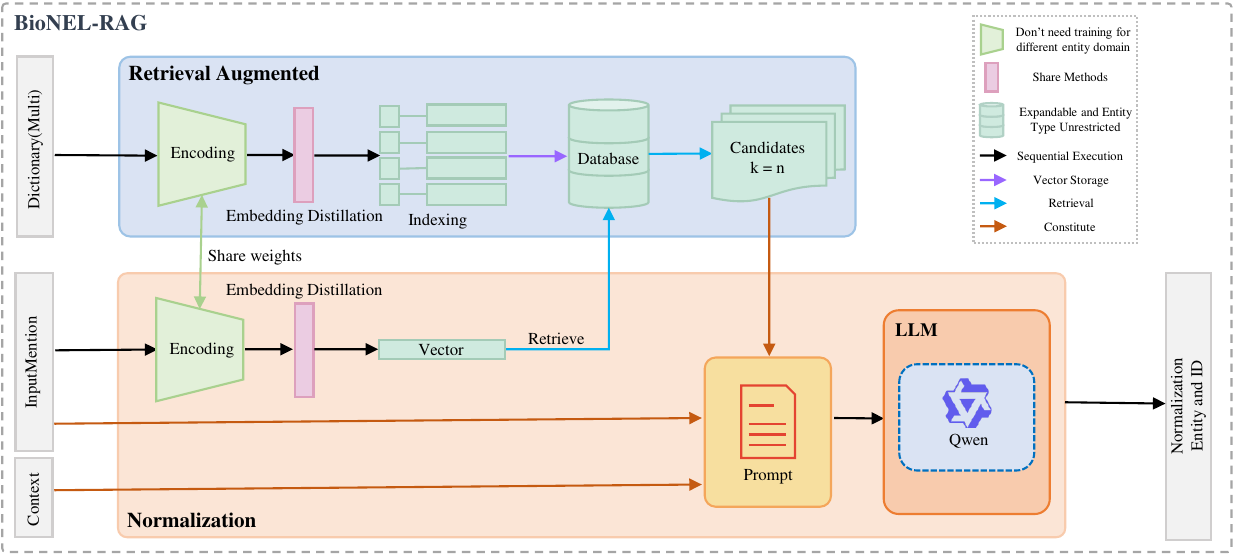} 
\caption{\textbf{The overall frameworks of BioNEL-RAG. BioNEL-RAG is composed of two core components: Retrieval Augmentation, which is responsible for handling the dictionary and building the knowledge base, and Normalization, which processes the mentions to be normalized.}}
\label{fig:bionel-rag}
\end{figure*}

%%%%%%%%%%%%%%%%%%%%%%%%%%%%%%%%%%%%%%%%%%%%%%%%%%%%%%%%%%%%

%%%%%%%%%%%%%%%%%%%%%%%%%%%%%%%%%%%%%%%%%%%%%%%%%%%%%%%%%%%%
%%%%%%%%%%%%%%%%%%%%%%%%%%%%%%%%%%%%%%%%%%%%%%%%%%%%%%%%%%%%
%%%%%%%%%%%%%%%%%%%%%%%%%%%%%%%%%%%%%%%%%%%%%%%%%%%%%%%%%%%%

\section{Materials and methods}
In this section, we present the implementation details of our method, and the overall workflow is illustrated in Fig.~\ref{fig:bionel-rag}. 

\subsection{Datasets}
This study focuses on the biomedical NEL, utilizing datasets that encompass multiple domains, including diseases, chemicals, and genes. The disease-related datasets include NCBI-Disease\cite{dougan2014ncbi} and BC5CDR-Disease\cite{li2016biocreative}, while the chemical-related dataset is BC5CDR-Chemical\cite{li2016biocreative}. For gene-related entities, the BC2GM-Gene\cite{smith2008overview} dataset is used. In addition to the aforementioned datasets, we also utilize the SMM4H-2017\cite{sarker2017overview} dataset, which contains content from social media texts. 

\subsection{BioNEL Retrieval-Augmented Generation}
In this study, we presents BioNEL-RAG, a novel, training-free framework that synergizes RAG with LLMs for biomedical entity linking (Fig.~\ref{fig:bionel-rag}). The architecture comprises two core processes: Retrieval Augmented and Normalization. For the Retrieval Augmented component, we populate a vector database by encoding a standard biomedical entity dictionary. Critically, our approach bypasses the conventional RAG text-chunking step, as the dictionary entries are inherently concise and structured. Each entry is transformed into a distilled high-dimensional vector via an Encoding and Embedding Distillation module before being indexed for efficient retrieval. The Normalization process begins by vectorizing an InputMention using the identical embedding and distillation module. This vector serves as a query to retrieve the top-k candidate entities from the knowledge base. To facilitate accurate disambiguation, these candidates, the InputMention, and its original sentence (Context) are synthesized into a structured prompt. This prompt enables the LLM to leverage contextual cues to output the final normalized entity.

%%%%%%%%%%%%%%%%%%%%%%%%%%%%%%%%%%%%%%%%%%%%%%%%%%%%%%%%%%%%

% \begin{figure}[htbp]
% \flushright
% \includegraphics[width=1.0\textwidth]{Figures/Normalization_Entity_output.pdf} % 
% \caption{\textbf{Overview of the process for candidate entity retrieval, generation, and the final output of normalized entities.}}
% \vspace{-1em}
% \label{fig:entities-selection}
% \end{figure}

%%%%%%%%%%%%%%%%%%%%%%%%%%%%%%%%%%%%%%%%%%%%%%%%%%%%%%%%%%%%

\subsubsection{Task Definition}

The task of the framework is defined as follows: we define the string of a mention as \textit{m}, and for each mention, there is a corresponding ID identifier, also known as a CUI. Theoretically, one ID can be associated with multiple mentions \textit{m}, reflecting a one-to-many relationship. Therefore, the core objective of this study is to accurately predict the $\text{ID}_m$ corresponding to the input mention \textit{m}.

\begin{equation}
\medmath{
    \text{ID}_{m_i}^* = \text{MapID}\left(\underset{M \in \mathcal{M}}{\text{argmax}} \, P(M \mid m_i)\right)
}
\label{MapID}
\end{equation}
Where \text{MapID(*)} is a mapping function that maps the most probable entity \textit{M} to its corresponding unique ID. $\mathcal{M}$ represents the set of all possible entity type identifiers. \textit{M} is a single entity type identifier from the set $\mathcal{M}$.

\subsubsection{Key Components} 

\textbf{Dictionary.} In this work, the dictionaries currently used to evaluate model performance are derived from publicly available datasets, and the processing approach for these datasets aligns with previous works\cite{sung-etal-2020-biomedical}.

\textbf{Embedding.} The task of this step is to transform the formatted textual content of the dictionary into vector representations. To generate informative dense representations of dictionary entries, we employ the SAPBERT-PubMedBERT\cite{liu2020self} encoder to capture their key features. The embedding of mention \textit{m} is calculated as the following equation:

\begin{equation}
\medmath{
    e_m = \text{PubMedBERT-SAPBERT}(m)[CLS] \in \mathbb{R}^h
}
\label{SapBERT}
\end{equation}
Where $e_m$ represents the embedding vector of mention \textit{m} outputted by the BERT model. [\textit{CLS}] denotes the special token that BERT-style models use to compute a single vector representation of the input. $\mathbb{R}^h$ denotes the hidden dimension of BERT-style models, where \textit{h} = 768.

\textbf{Embedding Distillation.}BERT-based models typically produce high-dimensional (e.g., 768-D) vectors. For concise texts like entity(\textit{C536227\texttt{||}cyclic neutropenia}), these vectors can introduce redundancy and increase computational overhead. To address this, we employ an Embedding Distillation method using Principal Component Analysis (PCA)\cite{abdi2010principal} to reduce dimensionality while preserving critical features. Our analysis determined that a dimensionality of 384 offers the optimal trade-off, and we therefore use 384-dimensional vectors for all subsequent experiments.

\begin{equation}
\medmath{
    e_d = \text{PCA}(e_m) = W^T e_m \in \mathbb{R}^{\mathrm{N \times k}}
}
\label{PCA}
\end{equation}
Where $e_m \in \mathbb{R}^{N \times h}$ represents the high-dimensional embedding matrix, $W \in \mathbb{R}^{h \times k}$ is the PCA projection matrix, and $e_d \in \mathbb{R}^{N \times k}$ is the reduced-dimensional embedding matrix, with $k = 384$. 

\textbf{Similarity Computation.} In this part, we employ the IndexFlatIP method for similarity search, which performs linear scanning using the inner product as the similarity metric. The principles of similarity calculation and retrieval are as follows:

\begin{equation}
\medmath{
    S_{\text{similarity}}(m, n) = e_m^d \cdot e_n^d
}
\label{similarity}
\end{equation}
Where $e_m^d$ and $e_n^d$ represent the reduced embeddings of a mention and its corresponding dictionary synonym, respectively.

\begin{equation}
\medmath{
    E_{\text{msimilar}} = \text{Search}(\text{IndexFlatIP}(S_{\text{similarity}}(m, n)), k)
}
\label{IndexFlatIP}
\end{equation}
Where $E_{\text{msimilar}}$ denotes the optimal set of top k candidate entities obtained from IndexFlatIP. 

%%%%%%%%%%%%%%%%%%%%%%%%%%%%%%%%%%%%%%%%%%%%%%%%%%%%%%%%%%%%

\begin{figure}[htb]
\centering
\includegraphics[width=0.48\textwidth]{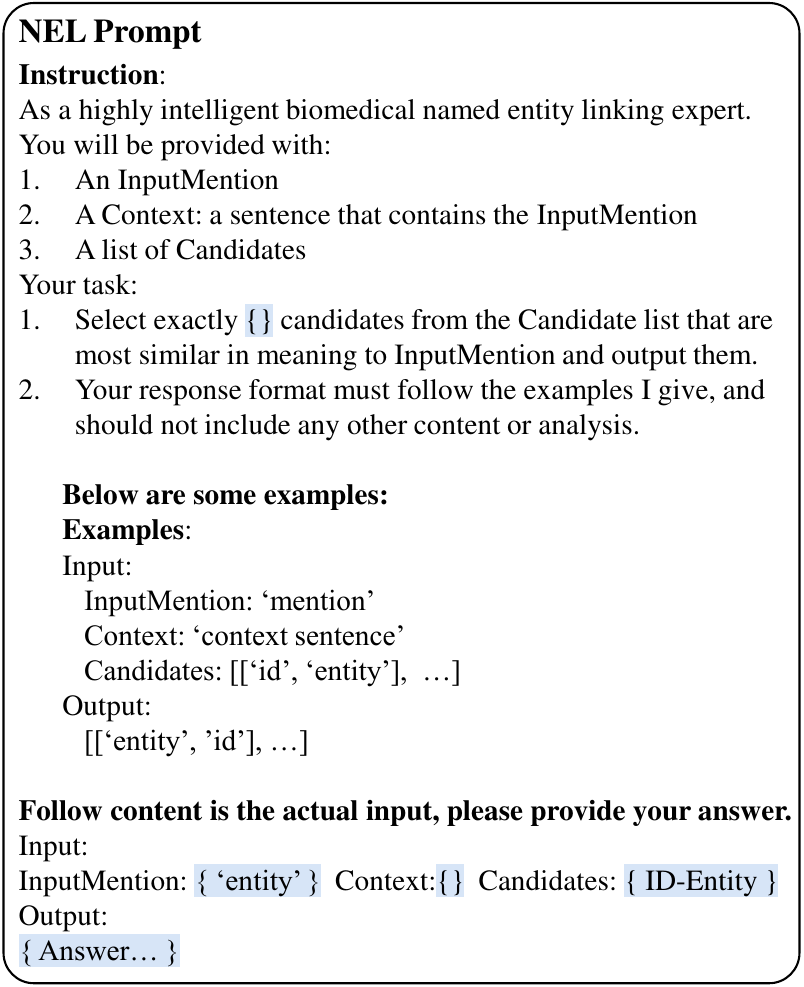} % 
\caption{\textbf{Prompt Design for biomedical named entity linking task by LLMs.}}
\vspace{-1em}
\label{fig:prompt}
\end{figure}

%%%%%%%%%%%%%%%%%%%%%%%%%%%%%%%%%%%%%%%%%%%%%%%%%%%%%%%%%%%%

\subsection{LLM and Prompt Design}

\subsubsection{Large Language Model}

For this framework, we selected Qwen3-8B\cite{yang2025qwen3}, released by Alibaba, for its superior IFEval performance and effective multitask learning compared to other models of similar size.

\subsubsection{Prompt Design}

Our prompt engineering is specifically designed for entity selection rather than generative synthesis. Unlike traditional RAG, which asks an LLM to generate an answer from retrieved text, our prompt instructs the LLM to choose the optimal entity from a candidate list.  To facilitate this, the prompt contains three key elements: the mention to be normalized, the retrieved candidate entities, and the original context sentence. Including the context is crucial: it provides the LLM with the necessary information to disambiguate the mention's meaning. For the detailed composition of the prompt, refer to Fig.~\ref{fig:prompt}.

%%%%%%%%%%%%%%%%%%%%%%%%%%%%%%%%%%%%%%%%%%%%%%%%%%%%%%%%%%%%
%%%%%%%%%%%%%%%%%%%%%%%%%%%%%%%%%%%%%%%%%%%%%%%%%%%%%%%%%%%%
%%%%%%%%%%%%%%%%%%%%%%%%%%%%%%%%%%%%%%%%%%%%%%%%%%%%%%%%%%%%

%%%%%%%%%%%%%%%%%%%%%%%%%%%%%%%%%%%%%%%%%%%%%%%%%%%%%%%%%%%%

\begin{table*}[htb]
\caption{Experimental results of models on three biomedical entity normalization datasets, with accuracy(Acc) serving as the evaluation metrics.}
\scriptsize % 调整表格字体
\begin{tabular*}{\textwidth}{@{\extracolsep{\fill}}ccccccccc}
\toprule
 & \multicolumn{2}{c}{{NCBI-Disease}} & \multicolumn{2}{c}{{BC5CDR-Disease}} & \multicolumn{2}{c}{{BC5CDR-Chemical}} & \multicolumn{2}{c}{{Average}} \\ 
 % \cline{2-3} \cline{4-5} \cline{6-7} \cline{8-9}
 \cmidrule{2-3}\cmidrule{4-5} \cmidrule{6-7} \cmidrule{8-9}
 & Acc@1 & Acc@5 & Acc@1 & Acc@5 & Acc@1 & Acc@5 & Acc@1 & Acc@5 \\
\midrule
Sieve-Based\cite{d2015sieve}                & 0.847 & - & 0.841 & - & 0.907 & - & 0.865 & - \\
Taggerone\cite{leaman2016taggerone}         & 0.874 & - & 0.889 & - & 0.941 & - & 0.901 & - \\
BNE\cite{phan2019robust}                    & 0.877 & - & 0.906 & - & 0.958 & - & 0.913 & - \\ 
BioBERT Ranking\cite{ji2020bert}               & 0.891 & - & 0.913 & - &  0.947 & - & 0.917 & - \\
BioSYN\cite{sung-etal-2020-biomedical}      & 0.911 & 0.939 & 0.924 & 0.958 & 0.963 & 0.972 & 0.933 & 0.956 \\
SAPBERT-PubMedBERT\cite{liu2020self} & 0.923 & 0.952 & 0.932 & 0.959 & 0.965 & 0.977 & 0.940 & 0.963 \\
SAPBERT-BioSYN\cite{liu2020self}     & 0.925 & 0.959 & 0.936 & 0.962 & 0.966 & \textbf{0.984} & 0.942 & \textbf{0.968} \\
GenBioEL\cite{yuan2022generative}           & 0.919 & \textbf{0.964} & 0.902 & 0.942 & 0.959 & 0.977 & 0.927 & 0.961 \\
Ours(non-EmbedDistil)                       & 0.931 & 0.959 & \textbf{0.937} & 0.963 & 0.967 & 0.980 & 0.945 & 0.967 \\
Ours(non-Context)                           & 0.931 & 0.958 & 0.932 & 0.962 & 0.966 & 0.980 & 0.943 & 0.967 \\
Ours                               & \textbf{0.932} & 0.958 & \textbf{0.937} & \textbf{0.964} & \textbf{0.969} & 0.981 & \textbf{0.946} & \textbf{0.968} \\ 
\bottomrule
\label{tab:main}
\end{tabular*}
\end{table*}

%%%%%%%%%%%%%%%%%%%%%%%%%%%%%%%%%%%%%%%%%%%%%%%%%%%%%%%%%%%%

\section{Experiments and Results}

To ensure a fair comparison across different methods, we adopt the top-$\textit{k}$ accuracy as the evaluation metric. We follow the definition of accuracy $\textit{Acc}$ introduced in BioSyn\cite{sung-etal-2020-biomedical}. %All experiments are conducted on a single 11GB GTX 1080Ti GPU, with the LLM running inference using 4-bit quantization.

\subsection{Model performance}
In this study, the experimental results of BioNEL-RAG presented in the following tables were obtained without training on the train sets of any specific datasets.

\subsubsection{Comparison of BioNEL-RAG with other models} 
Table~\ref{tab:main} presents the main performance comparison. Our training-free method, BioNEL-RAG, achieves results competitive with or superior to previous SOTA models. Notably, these SOTA baselines were fine-tuned on each respective dataset, whereas BioNEL-RAG requires no such training. Specifically, on the NCBI-Disease, BioNEL-RAG outperforms SAPBERT-BioSYN by 0.7\% and SAPBERT-PubMedBERT by 0.9\% in Acc@1. On both BC5CDR datasets, our method performs comparably to SAPBERT-BioSYN and consistently outperforms SAPBERT-PubMedBERT. 

%%%%%%%%%%%%%%%%%%%%%%%%%%%%%%%%%%%%%%%%%%%%%%%%%%%%%%%%%%%%

\begin{table}[htb]
\caption{Experimental results of learning-based models and BioNEL-RAG on unseen datasets.}
\scriptsize % 调整表格字体
\begin{tabular*}{\columnwidth}{@{\extracolsep{\fill}}ccccc}
\toprule
 & \multicolumn{2}{c}{{BC2GM-Gene}} & \multicolumn{2}{c}{{SMM4H-2017}} \\ 
  % \cline{2-3} \cline{4-5}
  \cmidrule{2-3} \cmidrule{4-5}
 & Acc@1 & Acc@5 & Acc@1 & Acc@5 \\ 
 \midrule
BioBERT ranking       & 0.753 & 0.821 & 0.339 & 0.423 \\
BioSYN                & 0.849 & 0.926 & 0.555 & 0.716 \\
SAPBERT-BioSYN       & 0.905 & 0.949 & 0.606 & 0.765 \\
Ours                  & \textbf{0.914} & 0.949 & \textbf{0.617} & \textbf{0.786} \\ 
\bottomrule
\label{tab:compare_unseen}
\end{tabular*}
\end{table}

%%%%%%%%%%%%%%%%%%%%%%%%%%%%%%%%%%%%%%%%%%%%%%%%%%%%%%%%%%%%

\subsubsection{Compare with models on unseen datasets}
To evaluate generalization, we test models on "unseen datasets," defined as those excluded from the training phase of the baseline models. This setup ensures a fair comparison of each model's performance on novel data. The specific models used for comparison are listed in Table~\ref{tab:compare_unseen}. The evaluation is conducted on the BC2GM-Gene and SMM4H-2017 datasets.

As shown in Table~\ref{tab:compare_unseen}, compared to the three models, our method demonstrates notable generalization capabilities on both datasets. Compared with SAPBERT-BioSYN, the improvements in Acc@1 on BC2GM-Gene and SMM4H-2017 are 0.9\% and 1.1\%, respectively, while the improvements in Acc@5 on SMM4H-2017 are 2.1\%.

\subsection{Ablation Studies}
To evaluate the efficacy of various components within the BioNEL-RAG framework, a series of ablation studies is conducted, as detailed in Table~\ref{tab:ablation}.

%%%%%%%%%%%%%%%%%%%%%%%%%%%%%%%%%%%%%%%%%%%%%%%%%%%%%%%%%%%%

\begin{table}[htb]
\caption{Ablation studies.}
\scriptsize % 调整表格字体
\begin{tabular*}{\columnwidth}{@{\extracolsep{\fill}}cccccc}
\toprule
\multirow{2}{*}{SAPBERT} & \multirow{2}{*}{MiniLM} & \multirow{2}{*}{EmbedDistil} & \multirow{2}{*}{Qwen} & \multicolumn{2}{c}{NCBI-Disease} \\ \cmidrule{5-6} 
 &  &  &  & Acc@1 & Acc@5 \\ 
\midrule
\checkmark &  &  &  & 0.920 & 0.952 \\
\checkmark &  & \checkmark &  & 0.919 & 0.946 \\
\checkmark &  &  & \checkmark & 0.931 & \textbf{0.959} \\
 & \checkmark & \checkmark & \checkmark & 0.819 & 0.921 \\
\checkmark &  & \checkmark & \checkmark & \textbf{0.932} & 0.958 \\ 
\bottomrule
\label{tab:ablation}
\end{tabular*}
\end{table}

%%%%%%%%%%%%%%%%%%%%%%%%%%%%%%%%%%%%%%%%%%%%%%%%%%%%%%%%%%%%

\subsubsection{Embedding Model}
To assess the impact of the embedding model on performance, we conduct an ablation study and compare the performance of two models: the general-purpose all-Mini-LM-v2\footnote{https://huggingface.co/sentence-transformers/all-MiniLM-L6-v2} and the domain-specific SAPBERT-PubMedBERT. Table~\ref{tab:ablation} (rows 4 and 5) show that the domain-specific model outperformed its generalist counterpart by a significant 11.3\% in Acc@1.

\subsubsection{Large Language Models}
To evaluate the contribution of the LLMs, we conducted an ablation study comparing our full framework against a variant with the LLM module removed.  As shown in rows 1 and 3 of Table~\ref{tab:ablation}, the model with the LLMs module outperforms the one without it by 2\% in both Acc@1 and Acc@5.

\subsubsection{Embedding Distillation}
This part evaluates the impact of the Embedding Distillation module. According to Table~\ref{tab:ablation}(rows 1 and 2), applying distillation not only avoids performance degradation but also yields slight improvements in both Acc@1 and Acc@5. Crucially, this module significantly reduces computational and storage costs.

\section{Conclusion}
In this paper, we propose BioNEL-RAG, a novel biomedical entity linking framework that integrates LLMs with RAG. Experimental results demonstrate that BioNEL-RAG achieves SOTA performance without requiring training, presenting a new paradigm for the field of NEL.
%Experimental results demonstrate that BioNEL-RAG achieves SOTA performance across multiple datasets and exhibits generalization on unseen data.

\section*{Acknowledgment}
This work is supported by funds from the National Key R\&D Program of China (Grant No.2022YFA1104800).

% \section*{References}
\bibliographystyle{IEEEtran}
\bibliography{reference}

\begin{thebibliography}{10}
\providecommand{\url}[1]{#1}
\csname url@samestyle\endcsname
\providecommand{\newblock}{\relax}
\providecommand{\bibinfo}[2]{#2}
\providecommand{\BIBentrySTDinterwordspacing}{\spaceskip=0pt\relax}
\providecommand{\BIBentryALTinterwordstretchfactor}{4}
\providecommand{\BIBentryALTinterwordspacing}{\spaceskip=\fontdimen2\font plus
\BIBentryALTinterwordstretchfactor\fontdimen3\font minus
  \fontdimen4\font\relax}
\providecommand{\BIBforeignlanguage}[2]{{%
\expandafter\ifx\csname l@#1\endcsname\relax
\typeout{** WARNING: IEEEtran.bst: No hyphenation pattern has been}%
\typeout{** loaded for the language `#1'. Using the pattern for}%
\typeout{** the default language instead.}%
\else
\language=\csname l@#1\endcsname
\fi
#2}}
\providecommand{\BIBdecl}{\relax}
\BIBdecl

\bibitem{sung-etal-2020-biomedical}
\BIBentryALTinterwordspacing
M.~Sung, H.~Jeon, J.~Lee, and J.~Kang, ``Biomedical entity representations with
  synonym marginalization,'' in \emph{Proceedings of the 58th Annual Meeting of
  the Association for Computational Linguistics}, D.~Jurafsky, J.~Chai,
  N.~Schluter, and J.~Tetreault, Eds.\hskip 1em plus 0.5em minus 0.4em\relax
  Online: Association for Computational Linguistics, Jul. 2020, pp. 3641--3650.
  [Online]. Available: \url{https://aclanthology.org/2020.acl-main.335}
\BIBentrySTDinterwordspacing

\bibitem{bodenreider2004unified}
O.~Bodenreider, ``The unified medical language system (umls): integrating
  biomedical terminology,'' \emph{Nucleic acids research}, vol.~32, no.
  suppl\_1, pp. D267--D270, 2004.

\bibitem{dougan2014ncbi}
R.~I. Do{\u{g}}an, R.~Leaman, and Z.~Lu, ``Ncbi disease corpus: a resource for
  disease name recognition and concept normalization,'' \emph{Journal of
  biomedical informatics}, vol.~47, pp. 1--10, 2014.

\bibitem{li2016biocreative}
J.~Li, Y.~Sun, R.~J. Johnson, D.~Sciaky, C.-H. Wei, R.~Leaman, A.~P. Davis,
  C.~J. Mattingly, T.~C. Wiegers, and Z.~Lu, ``Biocreative v cdr task corpus: a
  resource for chemical disease relation extraction,'' \emph{Database}, vol.
  2016, 2016.

\bibitem{smith2008overview}
L.~Smith, L.~K. Tanabe, R.~J.~n. Ando, C.-J. Kuo, I.-F. Chung, C.-N. Hsu, Y.-S.
  Lin, R.~Klinger, C.~M. Friedrich, K.~Ganchev \emph{et~al.}, ``Overview of
  biocreative ii gene mention recognition,'' \emph{Genome biology}, vol.~9, pp.
  1--19, 2008.

\bibitem{sarker2017overview}
A.~Sarker and G.~Gonzalez-Hernandez, ``Overview of the second social media
  mining for health (smm4h) shared tasks at amia 2017,'' \emph{Training},
  vol.~1, no. 10,822, p. 1239, 2017.

\bibitem{liu2020self}
F.~Liu, E.~Shareghi, Z.~Meng, M.~Basaldella, and N.~Collier, ``Self-alignment
  pretraining for biomedical entity representations,'' \emph{arXiv preprint
  arXiv:2010.11784}, 2020.

\bibitem{abdi2010principal}
H.~Abdi and L.~J. Williams, ``Principal component analysis,'' \emph{Wiley
  interdisciplinary reviews: computational statistics}, vol.~2, no.~4, pp.
  433--459, 2010.

\bibitem{yang2025qwen3}
A.~Yang, A.~Li, B.~Yang, B.~Zhang, B.~Hui, B.~Zheng, B.~Yu, C.~Gao, C.~Huang,
  C.~Lv \emph{et~al.}, ``Qwen3 technical report,'' \emph{arXiv preprint
  arXiv:2505.09388}, 2025.

\bibitem{d2015sieve}
J.~D’Souza and V.~Ng, ``Sieve-based entity linking for the biomedical
  domain,'' in \emph{Proceedings of the 53rd Annual Meeting of the Association
  for Computational Linguistics and the 7th International Joint Conference on
  Natural Language Processing (Volume 2: Short Papers)}, 2015, pp. 297--302.

\bibitem{leaman2016taggerone}
R.~Leaman and Z.~Lu, ``Taggerone: joint named entity recognition and
  normalization with semi-markov models,'' \emph{Bioinformatics}, vol.~32,
  no.~18, pp. 2839--2846, 2016.

\bibitem{phan2019robust}
M.~C. Phan, A.~Sun, and Y.~Tay, ``Robust representation learning of biomedical
  names,'' in \emph{Proceedings of the 57th Annual Meeting of the Association
  for Computational Linguistics}, 2019, pp. 3275--3285.

\bibitem{ji2020bert}
Z.~Ji, Q.~Wei, and H.~Xu, ``Bert-based ranking for biomedical entity
  normalization,'' \emph{AMIA Summits on Translational Science Proceedings},
  vol. 2020, p. 269, 2020.

\bibitem{yuan2022generative}
H.~Yuan, Z.~Yuan, and S.~Yu, ``Generative biomedical entity linking via
  knowledge base-guided pre-training and synonyms-aware fine-tuning,''
  \emph{arXiv preprint arXiv:2204.05164}, 2022.

\end{thebibliography}

\end{document}